%% file: main.tex
\documentclass[sigconf]{acmart}
\pdfoutput=1

\copyrightyear{2019}
\acmYear{2019}
\setcopyright{iw3c2w3}
\acmConference[WWW '19]{Proceedings of the 2019 World Wide Web Conference}{May 13--17, 2019}{San Francisco, CA, USA}
\acmBooktitle{Proceedings of the 2019 World Wide Web Conference (WWW '19), May 13--17, 2019, San Francisco, CA, USA}
\acmPrice{}
\acmDOI{10.1145/3308558.3313498}
\acmISBN{978-1-4503-6674-8/19/05}

\usepackage{balance}
\usepackage{graphicx}
\usepackage{amsmath}
\usepackage{subcaption}
\usepackage{hyperref}
\usepackage{array}
\usepackage[title]{appendix}
\usepackage{adjustbox}
\usepackage{multirow}
\usepackage{arydshln}
\usepackage{booktabs}
\usepackage{flushend}

\newcommand{\www}[1]{{#1}}

\newcommand{\commented}[1]{}

\newcommand\Tstrut{\rule{0pt}{1.1em}}       

\newcommand{\method}[1]{HAM$_{\text{#1}}$}
\newcommand{\paragraphHdTop}[1] {\noindent\textbf{#1}} 
\newcommand{\paragraphHd}[1] {\vspace{3pt}\noindent\textbf{#1}}

\newcommand{\squishlist}{
 \begin{list}{$\bullet$}
  { \setlength{\itemsep}{0pt}
     \setlength{\parsep}{1pt}
     \setlength{\topsep}{1pt}
     \setlength{\partopsep}{0pt}
     \setlength{\leftmargin}{1.5em}
     \setlength{\labelwidth}{1em}
     \setlength{\labelsep}{0.5em} } }

\newcommand{\squishend}{
  \end{list}  }
  
\newcommand{\sig}[1]{#1*}
\newcommand{\nsig}[1]{#1\phantom{*}}

\newcommand{\bnsig}[1]{\textbf{#1}\phantom{*}}

\begin{document}
 
\title{Listening between the Lines:\\
Learning Personal Attributes from Conversations}
\author{Anna Tigunova, Andrew Yates, Paramita Mirza, Gerhard Weikum}
\affiliation{%
  \institution{Max Planck Institute for Informatics}
  \city{Saarbr\"ucken}
  \country{Germany}
  }
\email{{tigunova, ayates, paramita, weikum}@mpi-inf.mpg.de}

\input{sections/abstract}

%
%

\ccsdesc[300]{Computing methodologies~Natural language processing}
\ccsdesc[500]{Computing methodologies~Information extraction}
\ccsdesc[500]{Computing methodologies~Supervised learning by classification}
\ccsdesc[500]{Computing methodologies~Neural networks}

\keywords{Information extraction; personal knowledge; user profiling; conversational text; neural networks; attention mechanisms}

\maketitle

\input{sections/intro}

\input{sections/related_work}

\input{sections/method}

\input{sections/data_acquisition}

\input{sections/experiments}

\input{sections/conclusion}
\balance

\newpage

\bibliographystyle{ACM-Reference-Format}
\balance 
\bibliography{bibliography}

\end{document}

%% file: sections/abstract.tex
\begin{abstract}
Open-domain dialogue agents must be able to converse about many topics while incorporating knowledge about the user into the conversation.
In this work we address the acquisition of such knowledge,
for personalization in downstream Web applications,
by extracting personal attributes from conversations.
This problem is more challenging than the established task of 
information extraction from scientific publications or Wikipedia articles, because dialogues
often give merely implicit cues about the speaker.
We propose methods for inferring personal attributes, such as
profession, age or family status, from conversations using deep learning.
Specifically, we propose several
\textit{Hidden Attribute Models}, which are 
neural networks 
leveraging 
attention mechanisms and embeddings. 
Our methods are trained on a per-predicate basis to output
rankings of object values for a given subject-predicate combination
(e.g., ranking the doctor and nurse professions high when speakers talk
about patients, emergency rooms, etc).
Experiments with various conversational texts including Reddit discussions, movie scripts and a collection of crowdsourced personal dialogues
demonstrate the viability of our methods and their superior performance compared
to state-of-the-art baselines.
\end{abstract}

%% file: sections/intro.tex
\section{Introduction}

\paragraphHd{Motivation and Problem\textnormal{:}}
While interest in dialogue agents has grown rapidly in recent years, creating agents capable of holding personalized conversations remains a challenge. 
For meaningful and diverse dialogues with a real person, a system should be able to infer knowledge about the person's background from her utterances. 
Consider the following example where $H$ stands for a human and $A$ for an agent:\vspace{0.1cm}\\
\hspace*{0.3cm}{H:} {\em What's the best place for having brekky?}\\
\hspace*{0.3cm}{A:} {\em The porridge at Bread and Cocoa is great.}\\
\www{
\hspace*{0.3cm}{H:} {\em Any suggestions for us and the kids later?} 
}\\
\hspace*{0.3cm}\phantom{H:} {\em We already visited the zoo.}\\
\www{
\hspace*{0.3cm}{A:} {\em There's the San Fransisco Dungeon,}\\
\hspace*{0.3cm}\phantom{H:} {\em an amusement ride with scary city history.}
}
\vspace{0.1cm}

\noindent From the word `brekky' in the first $H$ utterance, 
the system understands that the user is Australian
and may thus like porridge for breakfast.
However, the 
cue is missed that the user 
{is with pre-teen children (talking about kids and the zoo)},
and the resulting suggestion is inappropriate for young children. 
Instead, with awareness of this knowledge, a better reply could have been:\vspace{0.1cm}\\
\hspace*{0.3cm}{A:} {\em I bet the kids loved the sea lions, so you} \\ 
\hspace*{0.3cm}\phantom{A:} {\em should see the dolphins at 
\www{
Aquarium of the Bay.
}
}
\vspace{0.1cm}

\noindent A possible remedy to improve this situation is to include user information
into an end-to-end learning system for the dialogue agent. 
However, any user information would be bound to latent representations rather than explicit attributes.
Instead, we propose to capture such attributes and construct a 
{\em personal knowledge base (PKB)}
with this information, which will then be a distant source of background knowledge
for personalization in downstream applications such as
Web-based chatbots and agents in online forums.

The advantages of an explicit PKB are twofold:
it is an easily re-usable asset that benefits all kinds of applications
(rather than merely fixing the current discourse),
and it provides a convenient way of explaining the agent's statements to
the human whenever requested.
Constructing a PKB involves several key issues:
\squishlist
\item What is personalized knowledge about users?
\item How can this knowledge be inferred from the users' utterances in conversations?
\item How should such background knowledge be incorporated into a dialogue system?
\squishend

As for the first issue,
interesting facts about a user could be attributes (age, gender, occupation, etc), interests and hobbies, relationships to other people (family status, names of friends, etc) or sentiments towards certain topics or people. 
In this paper's experiments, we focus on crisp attributes like profession and gender.

\sloppy{
The second issue is concerned with information extraction from text.
However, prior works have mostly focused on well compre\-hens\-ible text genres
such as Wikipedia articles or news stories. 
These methods do not work as well when conversations are the input.
Compared to formal documents, dialogues are noisy, utterances are short, the language used is
colloquial, topics are diverse (including smalltalk), 
and information is often implicit (``between the lines'').
This paper addresses these problems by proposing methods that identify the terms that are informative for an attribute and leverage these terms to infer the attribute's value.
}

A detailed exploration of the third issue is beyond the scope of this paper, which focuses on inferring personal attributes. However, we do partially address the issue of integrating background knowledge by arguing that such information should be captured in an explicit PKB. In addition to being independent of downstream applications, an explicit PKB can provide transparency by allowing users to see what is known about them as well as giving users the opportunity to consent to this data being stored.


\paragraphHd{State of the Art and its Limitations\textnormal{:}}
Currently the most successful dialogue agents are task-oriented, 
for instance, supporting users with car navigation or delivery orders
(e.g., \cite{dial6,AAAI1816104}).
General-purpose chatbot-like agents show decent performance in
benchmarks (e.g., \cite{bot1,pers1,dial8}), 
but critically rely on sufficient training data and
tend to lack robustness when users behave in highly varying ways.
Very few approaches have considered incorporating explicit knowledge
on individual users, and these approaches have assumed that personal attributes
are explicitly mentioned in the text \cite{dial7,zhang2018personalizing,jing-kambhatla-roukos:2007:ACLMain}.

To illustrate that identifying explicit mentions of attributes is insufficient,
we developed an oracle to obtain an upper bound on the performance of pattern-based approaches, such as \cite{dial7}. This oracle, which is described in Section \ref{sec:baselines}, assumes that we have perfect pattern matching that correctly extracts an attribute value every time it is mentioned.
(When multiple attribute values are mentioned, we assume the oracle picks the correct one.)
This oracle routinely performs substantially worse than our proposed methods, demonstrating that extracting information from utterances requires \textit{listening between the lines}
(i.e., inferring the presence of attribute values that are never explicitly stated).

\www{
On the other hand, many efforts have considered the problem of profiling social media users in order to predict latent attributes such as age, gender, or regional origin (e.g., \cite{Rao:2010,burger:EMNLP11,Schwartz2013PersonalityGA,sap:EMNLP14,flekova:ACL16:long,kim:ACL17:short,vijayaraghavan:ACL17:short,bayot:MOD17,fabian2015privacy}). While social media posts and utterances are similar in that both are informal, the former can be associated with many non-textual features that are unavailable outside of the social media domain 
(e.g., social-network friends, likes, etc. and explicit self-portraits of users).
We consider several user profiling baselines that rely on only textual features and find that they do not perform well on our task of inferring attributes from 
conversational utterances.
}


\paragraphHd{Approach and Contributions\textnormal{:}}
We devise a neural architecture, called {\em Hidden Attribute Model (HAM)}, 
trained with
subject-predicate-object triples to predict objects on a per-predicate basis,
e.g., for a subject's profession or family status.
The underlying neural network learns to predict a scoring
of different objects (e.g., different professions) for
a given subject-predicate pair
by using attention within and across utterances to infer object values.
For example, as illustrated later in Table \ref{tab6}, our approach infers that a subject
who often uses terms like \textit{theory}, \textit{mathematical}, and \textit{species} is likely to be a \textit{scientist}, while a subject who uses terms like \textit{senate}, \textit{reporters}, and \textit{president} may be a \textit{politician}.

Our salient contributions are the following: 
\squishlist
\item a viable method for learning personal attributes from
conversations, based on neural networks with novel ways of
leveraging attention mechanisms and embeddings,
\item a data resource of dialogues from movie scripts, with
ground truth for professions, genders, and ages of movie characters,
\item \www{a data resource of dialogue-like Reddit posts,
with ground truth for the authors' professions, genders, ages, and family statuses}
\item an extensive experimental evaluation of various methods on
\www{Reddit, }
movie script dialogues, and crowdsourced personalized conversations (PersonaChat),
\item \www{
an experimental evaluation on the transfer learning approach: 
leveraging
ample data from
user-generated social media text (Reddit) for inferring users' latent attributes from data-scarce speech-based dialogues (movie scripts and PersonaChat).
}
\squishend


%% file: sections/related_work.tex
\section{Related work}

\paragraphHd{Personal Knowledge from Dialogues:} Awareness of background knowledge about users is important 
in both goal-oriented dialogues \cite{Chen2016,pers2} and chitchat settings \cite{pers1,zhang2018personalizing}. 
Such personal knowledge may range from users' intents and goals \cite{Chen2016,wakabayashi-EtAl:2016:EMNLP2016} to users' profiles, with attributes such as home location, gender, age, etc. \cite{pers2,pers1,AIIDElin11}, as well as users' preferences \cite{AAAI1816104,zhang2018personalizing}. 

Prior works on incorporating background knowledge 
model \textit{personas} 
and enhance dialogues
by injecting personas into the agent's response-generating model \cite{pers1,AIIDElin11,zhang2018personalizing} 
or by adapting the agent's speech style \cite{pers2}. 
These approaches typically use latent representations of 
user information.
In contrast, there is not much work on 
capturing explicit knowledge based on user utterances. 
\citeauthor{pers1} 
encode implicit speaker-specific information via distributed embeddings that can be used to cluster users along some traits (e.g.,
age or country) \cite{pers1}.
\citeauthor{zhang2018personalizing} 
describe
a preliminary study on predicting simple speaker profiles 
from a set of dialogue utterances \cite{zhang2018personalizing}. 
The most related prior work
by \citeauthor{garera-yarowsky:2009:ACLIJCNLP} captures latent biographic attributes 
via SVM models from conversation transcripts and email communication, taking into account \emph{contextual features} such as partner effect and sociolinguistic features (e.g., \% of ``yeah'' occurences), in addition to n-grams \cite{garera-yarowsky:2009:ACLIJCNLP}.

Several works explore information extraction from conversational text to build a personal knowledge base of speakers given their utterances, such as extracting \textit{profession}: \textit{software engineer} and \textit{ employment\_history}: \textit{Microsoft} from \textit{``I work for Microsoft as a software engineer''}, using maximum-entropy classifiers \cite{jing-kambhatla-roukos:2007:ACLMain} and sequence-tagging CRFs \cite{dial7}.
\commented{
The most related prior work
uses a sequence-tagging CRF to construct a personal knowledge base from 
speakers' utterances, such as extracting \textit{profession}: \textit{software engineer} and \textit{ employment\_history}: \textit{Microsoft} from \textit{``I work for Microsoft as a software engineer''} \cite{dial7}. 
}
However, this approach relies on user attributes being {\em explicitly mentioned} in the utterances.
In contrast, our method can infer attribute values from implicit cues,
e.g., \textit{``I write product code in Redmond.''}

\www{
\paragraphHd{Social Media User Profiling:} The rapid growth of social media 
has led to a massive volume of user-generated informal text, which sometimes mimics conversational utterances.
A great deal of work has been dedicated to automatically identify latent demographic features of online users, including age and gender \cite{Rao:2010,burger:EMNLP11,Schwartz2013PersonalityGA,sap:EMNLP14,flekova:ACL16:long,kim:ACL17:short,vijayaraghavan:ACL17:short,bayot:MOD17,fabian2015privacy}, political orientation and ethnicity \cite{Rao:2010,pennacchiotti:ICWSM11,pietro:ACL17:long,pietro:COLING18,vijayaraghavan:ACL17:short}, regional origin \cite{Rao:2010,fabian2015privacy}, personality \cite{Schwartz2013PersonalityGA,gjurkovic-EtAl:2018}, as well as occupational class that can be mapped to income \cite{pietro:ACL15,flekova:ACL16:short}.
Most of these works focus on user-generated content from Twitter, with a few exceptions that explore Facebook \cite{Schwartz2013PersonalityGA,sap:EMNLP14} or Reddit \cite{fabian2015privacy,gjurkovic-EtAl:2018} posts.
}

\www{
Most existing studies on social media to capture users' latent attributes rely on classification over hand-crafted features such as word/character n-grams \cite{Rao:2010,burger:EMNLP11,basile:2017}, Linguistic Inquiry and Word Count (LIWC) \cite{pennebaker2001linguistic} categories \cite{pietro:ACL17:long,pietro:COLING18,gjurkovic-EtAl:2018}, topic distributions \cite{pennacchiotti:ICWSM11,pietro:ACL15,flekova:ACL16:long} and sentiment/emotion labels of words derived from existing emotion lexicon \cite{pennacchiotti:ICWSM11,pietro:ACL17:long,pietro:COLING18,gjurkovic-EtAl:2018}. 
The best performing system \cite{basile:2017} in the shared task on author profiling organized by the CLEF PAN lab \cite{stein:2017l,stein:2017o} utilizes a linear SVM and word/character n-gram features.
}

\www{
There have been only limited efforts to identify attributes of social media users using neural network approaches. 
\citeauthor{bayot:MOD17} 
explore the use of Convolutional Neural Networks (CNNs) together with word2vec embeddings to perform user profiling (age and gender) of English and Spanish tweets \cite{bayot:MOD17}. \citeauthor{kim:ACL17:short} 
employ Graph Recursive Neural Networks (GRNNs) to infer demographic characteristics of users \cite{kim:ACL17:short}. \citeauthor{vijayaraghavan:ACL17:short} 
exploit attention-based models to identify demographic attributes of Twitter users given multi-modal features extracted
from users' profiles (e.g., name, profile picture, and description), social network, and tweets \cite{vijayaraghavan:ACL17:short}. 
}

\www{
The vast majority of these works rely on features specific to social media  such as hashtags, users' profile descriptions and social network structure, with only \cite{pietro:ACL15,bayot:MOD17,basile:2017} inferring users' latent attributes based solely on user-generated text. In our evaluation we consider these three methods as baselines.
}

\paragraphHd{Neural Models with Attention:} Recently, neural models enhanced with \textit{attention} mechanisms have boosted
the results of various NLP tasks \cite{atten1,atten9,atten2}, particularly 
in neural conversation models for generating responses \cite{atten8,atten7}
and in predicting user intents \cite{Chen2016}.
While our attention approach bears some similarity to \cite{atten8} in that both consider attention on utterances, both our approach's architecture and our use case differ substantially.
The role of attention weights has been studied for various neural models, including feed-forward networks \cite{vaswani2017attention}, CNNs, \cite{atten4} and RNNs \cite{bahdanau2014neural}.

%% file: sections/method.tex
\section{Methodology}
\label{sec:method}

\begin{figure}[t]
\centering
\texttt{\textmd{\footnotesize{
EDWARDS\\ 
Put down the gun and 
put your hands on the counter!\\
\vspace{4pt}
KAY\\ 
I warned him.\\
\vspace{4pt}
EDWARDS\\ 
Drop the weapon!\\
\vspace{4pt}
KAY\\ 
You warned him.\\
\vspace{4pt}
EDWARDS\\ 
You are under arrest. \\ 
You have the right to remain silent.\\
}}}
\caption{An excerpt from \textit{Men in Black} (1997).}
\label{fig:movie-script}
\end{figure}

In this section we propose Hidden Attribute Models (HAMs) 
for ranking object values given a predicate and a sequence of utterances made by a subject.
For example, given the movie script excerpt shown in Figure \ref{fig:movie-script}, 
the \textit{profession} predicate, and the subject \textit{Edwards},
\textit{policeman} should be ranked as the subject's most likely profession.

More formally, given a subject $S$ and a predicate $P$, our goal is to predict a probability distribution over object values $O$ for the predicate based on the subject's utterances from a dialogue corpus (e.g., a movie script). Each subject $S$ is associated with a sequence of $N$ utterances $[U_{1}, U_{2}, ..., U_{N}]$ containing $M$ terms each, $U_1 = [U_{1,1}, U_{1,2}, ..., U_{1,M}]$. Each term $U_{n,m}$ is represented as a $d$-dimensional word embedding.

HAMs
can be described in terms of three functions and their outputs:
\begin{enumerate}
\item $f_{utter}$ creates a representation $R^{utter}_n$  of the $nth$ utterance given the terms in the utterance:
\begin{equation}
R^{utter}_n = f_{utter}(U_{n,1}, U_{n,2}, ..., U_{n,M})
\end{equation}
\item $f_{subj}$ creates a subject representation $R^{subj}$ given the sequence of utterance representations:
\begin{equation}
R^{subj} = f_{subj}(R^{utter}_1, R^{utter}_2, ..., R^{utter}_N)
\end{equation}
\item $f_{obj}$ outputs a probability distribution over object values $O$ given the subject representation:
\begin{equation}
O = f_{obj}(R^{subj})
\end{equation}
\end{enumerate}

\noindent
In the following sections we describe our proposed 
HAMs
by instantiating these functions.

\subsection{Hidden Attribute Models}

\paragraphHdTop{\method{avg}} illustrates the most straightforward way to combine embeddings and utterances.
In this model,
\begin{equation}
avg(X) = \sum_{i=1}^{|X|} X_i
\end{equation}
serves as both $f_{utter} $ and $f_{subj}$; the $n$th utterance representation $R^{utter}_n$ is created by averaging the terms in the $n$th utterance and the subject representation $R^{subj}$ is created by averaging the $N$ utterance representations together. Two stacked fully connected layers serve as the function $f_{obj}$,
\begin{equation}
FC(x) = \sigma(Wx+b)
\end{equation}
where $\sigma$ is an activation function and $W$ and $b$ are learned weights.
The full \method{avg} model is then
\begin{equation}
R^{utter}_n = avg(U_n)
\end{equation}
\begin{equation}
R^{subj} = avg(R^{utter})
\end{equation}
\begin{equation}
O = FC_1(FC_2(R^{subj}))
\end{equation}
where $FC_2$ uses a sigmoid activation and $FC_1$ uses a softmax activation function in order to predict a probability distribution over object values.

\paragraphHd{\method{2attn}} extends \method{avg} with two attention mechanisms allowing the model to learn which terms and utterances to focus on for the given predicate. In this model the utterance representations and subject representations are computed using attention-weighted averages,
\begin{equation}
attn{\text -}avg(X, \alpha) = \sum_{i=1}^{|X|} X_i \alpha_i
\end{equation}
with the attention weights calculated over utterance terms and utterance representations, respectively.
That is, $f_{utter}(X)=attn{\text -}avg(X, \alpha^{term})$ and $f_{subj}(X)=attn{\text -}avg(X, \alpha^{utter})$, where
the attention weights for each term in an utterance $U_i$ are calculated as
\begin{equation} \label{eq:attn1}
w^{term}_i = \sigma(W^{term} U_i + b^{term})
\end{equation}
\begin{equation} \label{eq:attn2}
\alpha^{term}_{i,j} = \frac{exp(w^{term}_{i,j})}{\sum_j exp(w^{term}_{i,j})}
\end{equation}
and the utterance representation weights $\alpha^{utter}$ are calculated analogously over $R^{utter}$.
Given these attention weights, the \method{2attn} model is
\begin{equation}
R^{utter}_n = attn{\text -}avg(U_n, \alpha^{term})
\end{equation}
\begin{equation}
R^{subj} = attn{\text -}avg(R^{utter}, \alpha^{utter})
\end{equation}
\begin{equation}
O = FC(R^{subj})
\end{equation}
where $f_{obj}$ function $FC$ uses a softmax activation function as in the previous model.

\paragraphHd{\method{CNN}} considers n-grams when building utterance representations, unlike both previous models that treat each utterance as a bag of words. In this model $f_{utter}$ is implemented with a text classification CNN \cite{cnn} with a ReLU activation function and $k$-max pooling across utterance terms (i.e., each filter's top $k$ values are kept).
A second $k$-max pooling operation across utterance representations serves as $f_{subj}$.
As in the previous model, a single fully connected layer with a softmax activation function serves as $f_{obj}$.

\paragraphHd{\method{CNN-attn}} extends \method{CNN} by using attention to combine utterance representations into the subject representation.
This mirrors the approach used by \method{2attn}, with $f_{subj}=attn{\text -}avg(X, \alpha^{utter})$ and $\alpha^{utter}$ computed using equations \ref{eq:attn1} and \ref{eq:attn2} as before.
This model uses the same $f_{utter}$ and $f_{obj}$ as \method{CNN}. That is, utterance representations are produced using a CNN with $k$-max pooling, and a single fully connected layer produces the model's output.

\subsection{Training}
All 
HAMs
were trained with gradient descent to minimize a categorical cross-entropy loss. 
We use the Adam optimizer \cite{adam} with its default values and apply an L2 weight decay (2e-7) to the loss.

%% file: sections/data_acquisition.tex
\section{Data acquisition and processing}
\label{sec:data}

\paragraphHd{MovieChAtt dataset\textnormal{.}} Following prior work on personalized dialogue systems, we explore the applicability of fictional dialogues from TV or movie scripts to approximate real-life conversations \cite{pers1,AIIDElin11}.
Specifically, we compile a Movie Character Attributes (MovieChAtt) dataset consisting of characters' utterances and the characters' attributes (e.g., profession).

To create MovieChAtt, we labeled a subset of characters in the Cornell Movie-Dialogs Corpus\footnote{\href{https://www.cs.cornell.edu/~cristian/Cornell_Movie-Dialogs_Corpus.html}{http://www.cs.cornell.edu/~cristian/Cornell\_Movie-Dialogs\_Corpus.html}} \cite{Danescu-Niculescu-Mizil+Lee:11a} of 617 movie scripts.
From each movie, we derive a sequence of utterances for each character, excluding characters who have less than 20 lines in the movie.
Each utterance is represented as a sequence of words, excluding stop words, the 1,000 most common first names\footnote{Removed to prevent overfitting. \href{https://catalog.data.gov/dataset/baby-names-from-social-security-card-applications-national-level-data}{http://catalog.data.gov/dataset/baby-names-from-social-security-card-applications-national-level-data}}, and words that occur in fewer than four different movies.
The latter two types of words are excluded in order to prevent the model from relying on movie-specific or character-specific signals that will not generalize.

We extracted characters' \textit{gender} and \textit{age} attributes by associating characters with their entries in the Internet Movie Database (IMDb) and extracting the corresponding actor or actress' attributes at the time the movie was filmed.
This yielded 1,963 characters labeled with their genders and 4,548 characters labeled with their ages.
We discretized the age attribute into the following ranges:
(i) 0--13: \textit{child}, (ii) 14--23: \textit{teenager}, (iii) 24--45: \textit{adult}, (iv) 46--65: \textit{middle-aged} and (v) 66--100: \textit{senior}.
In our data the distribution of age categories is highly imbalanced, with \textit{adult} characters dominating the dataset (58.7\%) and \textit{child} being the smallest category (1.7\%).

To obtain the ground-truth labels of characters' \textit{profession} attributes, we conducted a Mechanical Turk (MTurk) crowdsourcing task to annotate 517 of the movies in our corpus.
Workers were asked to indicate the professions of characters in a movie given
the movie's Wikipedia article.
Workers were instructed to select professions from a general predefined list if possible (e.g., doctor, engineer, military personnel), and to enter a new profession label when necessary.
We manually defined and refined the list of professions based on several iterations of MTurk studies to ensure 
high coverage
and to reduce ambiguity in the options (e.g., \textit{journalist} vs \textit{reporter}).
We also included non-occupational ``professions'' that often occur in movies,
such as \textit{child} and \textit{criminal}.

Fleiss' kappa for the crowdworkers' inter-annotator agreement is $0.47$.
Disagreement was oftentimes caused by one character having multiple professions (Batman is both a \textit{superhero} and a \textit{businessman}), or a change of professions in the storyline (from \textit{banker} to \textit{unemployed}).
We kept only characters for which at least 2 out of 3 workers agreed on their profession,
which yielded 1405 characters labeled with 43 distinct professions.
The highly imbalanced distribution of professions reflects the bias in our movie dataset, which features more \textit{criminals} and \textit{detectives} than \textit{waiters} or \textit{engineers}.

\paragraphHd{PersonaChat dataset\textnormal{.}} We also explore the robustness of our models using the PersonaChat corpus\footnote{\href{http://convai.io/\#personachat-convai2-dataset}{http://convai.io/\#personachat-convai2-dataset}} \cite{zhang2018personalizing},
which consists of conversations collected via Mechanical Turk.
Workers were given persona descriptions consisting of 5-sentence-long descriptions (e.g., ``I am an artist'', ``I like to build model spaceships'') and asked to incorporate these facts into a short conversation (up to 8 turns) with another worker.
We split these conversations by persona, yielding a sequence of 3 or 4 utterances for each persona in a conversation.

We automatically labeled personas with \textit{profession} and \textit{gender} attributes by looking in the persona description for patterns ``I am/I'm a(n) $\langle$\textit{term}$\rangle$'', where 
$\langle$\textit{term}$\rangle$ is either a profession label
or a gender-indicating noun (woman, uncle, mother, etc).
We manually labeled persons with \textit{family status} by identifying persona
descriptions containing related words (single, married, lover, etc) and labeling the corresponding persona as single or not single.
Overall, we collected 1,147 personas labeled with profession, 1,316 with gender, and 2,302 labeled with family status.

\begin{table}[t!]
\centering
\small
\begin{tabular}{@{}l@{\hskip 0.5\tabcolsep}l@{}}
\toprule
\textbf{attribute} & \textbf{pattern} \\
\midrule
profession & ``(I|i) (am|'m)''+{[}one of the profession names{]} \\ [1.0ex]
gender & ``(I|i) (am|'m)''+{[}words, indicating gender (\textit{lady}, \textit{father}, etc.){]} \\ [1.0ex]
\multirow{2}{*}{age} & ``(I|i) (am|'m)''+number (5-100)+``years old'' \\ 
 & ``(I|i) (am|'m|was) born in''+number (1920-2013) \\ [1.0ex]
\multirow{2}{*}{family status} & ``(I|i) (am|'m)''+{[}\textit{married}, \textit{divorced}, \textit{single}, ... {]} \\
 & ``(M|m)y''|``(I|i) have a''+{[}\textit{wife}, \textit{boyfriend}, ...{]} \\
\bottomrule
\end{tabular}
\caption{Patterns for extracting ground truth labels from Reddit posts.}
\label{pat_table}
\end{table}

\www{
\paragraphHd{Reddit dataset\textnormal{.}}
As a proxy for dialogues we used discussions from Reddit online forums. We used a publicly available crawl\footnote{ https://files.pushshift.io/reddit/ } spanning the years from 2006 to 2017. Specifically, we tapped into two subforums on Reddit: ``iama'' where anyone can ask questions to a particular person, 
and ``askreddit'' with more general conversations. In selecting these subforums we followed two criteria: 1) they are not concerned with fictional topics (e.g. computer games) and 2) they are not too topic-specific, 
as this could heavily bias the classification of user attributes.

To create ground-truth labels for users, we searched for posts that matched specific patterns. The list of patterns for all four attributes is given in Table \ref{pat_table}. For the case of \emph{profession}, we created a list of synonyms and subprofessions for better coverage.
For example, the list for \textit{businessperson} includes \textit{entrepreneur}, \textit{merchant}, \textit{trader}, etc.

We selected only posts of length between 10 and 100 words by users who have between 20 and 100 posts.
Also, we removed users who claim multiple values for the
same attribute, as we allow only for single-label classification. 
To further increase data quality, we rated users by language style, to give preference to those whose posts sound more like the utterances in a dialogue. This rating was computed by
the fraction of a user's posts that contain personal pronouns, 
as pronouns are known to be abundant in dialogue data.

The test set, disjoint from training data and created in the same manner, was further checked by manual annotators, because the above mentioned patterns may also produce false positives,
for example, the  wrong profession from utterances such as ``they think I am a doctor'' or ``I dreamed  I am an astronout'', or the wrong age and family status
from ``I am 10 years old boy's mother'' or
``I am a single child''.
The final Reddit test set consists of approximately 400 users per predicate.

} 

\paragraphHd{Limitations\textnormal{.}} The predicates we consider are limited by the nature of our datasets. We do not consider the \textit{family status} predicate for MovieChAtt, because the necessary information is often not easily available from Wikipedia articles. Similarly, we do not consider the \textit{age} predicate on the PersonaChat dataset, because this attribute is not easily found in the persona descriptions. More generally, all users in our datasets are labeled with exactly one attribute value for each predicate. In a real setting it may be impossible to infer any attribute value for some users, whereas other users may have multiple correct values. We leave such scenarios for future work.

Our code and dataset annotations are available online.\footnote{\url{https://github.com/Anna146/HiddenAttributeModels}}\footnote{\url{https://mpi-inf.mpg.de/departments/databases-and-information-systems/research/personal-knowledge-base}}

%% file: sections/experiments.tex
\section{Experimental setup}

\subsection{Data}
We randomly split the MovieChAtt and PersonaChat 
datasets into training (90\%) and testing (10\%) sets.
The Reddit test set consists of approximately 400 users per predicate whose label extractions were manually verified by annotators.
We tuned models' hyperparameters by performing a grid search with 10-fold cross validation on the training set.

For the binary predicates family status and gender, we balanced the number of subjects in each class.
For the multi-valued profession and age predicates, which have very skewed distributions, we did not balance
the number of subjects in the test set. During training we performed downsampling to reduce the imbalance;
each batch consisted of an equal amount (set to $3$) of training samples per class, and these samples were drawn randomly for each batch. This both removes the class imbalance in training data and ensures that ultimately the model sees all instances during training, regardless of the class size.

Note that all three datasets are somewhat biased
regarding the attribute values and not representative
for real-life applications. For example,
the professions in MovieChAtt are dominated by
the themes of entertaining fiction, 
and gender distributions are not realistic either.
The data simply provides a diverse range of
inputs for fair comparison across different
extraction methods.

\setlength\dashlinedash{0.2pt}
\setlength\dashlinegap{1.5pt}
\setlength\arrayrulewidth{0.3pt}

\newcolumntype{L}{>{$}l<{$}}
\newcolumntype{C}{>{$}c<{$}}
\newcolumntype{R}{>{$}r<{$}}

\subsection{Baselines}
\label{sec:baselines}
We consider the following baselines to compare our approach against.

\paragraphHd{Pattern matching oracle}.
Assuming that we have a perfect sequence tagging model (e.g., \cite{dial7}) that extracts a correct attribute value every time one appears in an utterance, we can determine the upper-bound performance of such sequence tagging approaches. Note that, as mentioned in the introduction, this type of model assumes attribute values explicitly appear in the text.
In order to avoid vocabulary mismatches between our attribute value lists and the attribute values explicitly mentioned in the data, we augment our attribute values with synonyms identified in the data
(e.g., we add terms like `soldier' and `sergeant' as synonyms of the 
the \textit{profession} attribute's value `military personnel').
For both MRR and accuracy, a subject receives a score of 1 if the correct attribute value appears in any one of a subject's utterances. If the correct attribute value never appears, we assume the model returns a random ordering of attribute values and use the expectation over this list (i.e., given $|V|$ attribute values, the subject receives a score of $\frac{1}{0.5|V|}$ for MRR and $\frac{1}{|V|}$ for accuracy). This oracle method does not provide class confidence scores, so we do not report AUROC with this method.

\paragraphHd{Embedding similarity}.
Given an utterance representation created by averaging the embeddings of the words within the utterance,
we compute the cosine similarity between this representation and the embeddings of each attribute value.

\paragraphHd{Logistic regression}.
Given an averaged utterance representation (as used with embedding similarity),
we apply a multinomial logistic regression model to classify the representation as one of the possible attribute values. This model obtains a ranking by ordering on the per-class probabilities of its output.

\paragraphHd{Multilayer Perceptron (MLP)}. Given an averaged utterance representation (as used with embedding similarity), we apply an MLP with one hidden layer of size 100 to classify the utterance representation as one of the possible attribute values. This model can be used to obtain a ranking by considering the per-class probabilities.

\vspace{5pt}The embedding similarity, logistic regression, and MLP baselines are distantly supervised, because the subject's labels are applied to each of the subject's utterances. While this is necessary because the baselines do not incorporate the notion of a subject with multiple utterances, it results in noisy labels because it is unlikely that every utterance will contain information about the subject's attributes.
We address this issue by using a window of $k = 4$ (determined by a grid search) concatenated utterances as input to each of these methods. With these distantly supervised models, the label prediction scores are summed across all utterances for a single subject and then ranked.

\www{
\paragraphHd{CNN} \cite{bayot:MOD17}. We consider the Convolutional Neural Network (CNN) architecture proposed by \citeauthor{bayot:MOD17} for the task of predititng the age and gender of Twitter users. This approach is a simpler variant of \method{CNN} in which $f_{utter}$ is implemented with a \emph{tanh} activation function and max pooling (i.e., $k=1$) and $f_{obj}$ is a fully connected layer with dropout ($p=0.5$) and a softmax activation function.
The CNN is applied to individual utterances and the majority classification label is used for the user, which differs from the in-model aggregation performed by \method{CNN}.

  \paragraphHd{New Groningen Author-profiling Model (N-GrAM)} \cite{basile:2017}. Following the best performing system at CLEF 2017's PAN shared task on \textit{author profiling} \cite{stein:2017o}, we implemented a classification model using a linear Support Vector Machine (SVM) that utilizes the following features: character n-grams with $n=3,4,5$, and term unigrams and bigrams with sublinear tf-idf weighting.

\paragraphHd{Neural Clusters (W2V-C)} \cite{pietro:ACL15}. We consider the best classification model reported by \citeauthor{pietro:ACL15} for predicting the occupational class of Twitter users, which is a Gaussian Process (GP) classifier with \textit{neural clusters} (W2V-C) as features. Neural clusters were obtained by applying spectral clustering on a word similarity matrix (via cosine similarity of pre-trained word embeddings) to obtain $n=200$ word clusters. Each post's feature vector is then represented as the ratio of words from each cluster.

\vspace{5pt}For both N-GrAM and W2V-C baselines, flattened representations of the subject's utterances are used. That is, the model's input is a concatenation of all of a given user's utterances.
}

\subsection{Hyperparameters}
Hyperparameters were chosen by grid search using ten-fold cross validation on the training set.
Due to the limited amount of data, we found a minibatch size of 4 users performed best on MovieChAtt and PersonaChat.
All models were trained with a minibatch size of 32 on Reddit.
We used 300-dimensional word2vec embeddings pre-trained on the Google News corpus \cite{embed1} to represent the terms in utterances.
We set the number of utterances per character $N=40$ and the number of terms per utterance $M=40$, and truncated or zero padded the sequences as needed.

\squishlist
\item \textbf{\method{avg}} uses a hidden layer of size 100 with the sigmoid activation function. The model was trained for 30 epochs.
\item \textbf{\method{CNN}} uses 178 kernels of size 2 and k-max pooling with $k=5$. The model was trained for 40 epochs.
\item \textbf{\method{2attn}} uses a sigmoid activation with both attention layers and with the prediction layer.
The model was trained for 150 epochs.
\item \textbf{\method{CNN-attn}} uses 128 kernels of size 2. The model was trained for 50 epochs.
\squishend

 \subsection{Evaluation metrics}
\label{metric}

For binary predicates (\textit{gender} and \textit{family status}), we report models' performance in terms of accuracy. Due to the difficulty of performing classification over many attribute values, a ranking metric is more informative for the multi-valued predicates (\textit{profession} and \textit{age category}). We report Mean Reciprocal Rank for these predicates.

Mean Reciprocal Rank (MRR) measures the position of the correct answer in a ranked list of attribute values provided by the model. We obtain a ranking of attribute values for a movie character or a persona by considering the attribute value probabilities output by a model.
Given one ranked list of attribute values per character, MRR is computed by
determining the reciprocal rank of the correct attribute value in each list, and then taking the average of these reciprocal ranks.
We report both \textit{macro MRR}, in which we calculate a reciprocal rank for each attribute value before averaging, and \textit{micro MRR}, which averages across each subject's reciprocal rank.

Area under the ROC Curve (AUROC) measures the performance of a model as a function of its true positive rate and false positive rate at varying score thresholds. We report micro AUROC for all predicates. For multi-valued predicates, we binarize the labels in a one-vs-all fashion.

\begin{table*}[th!]
\centering
\small
\begin{tabular}{@{}lcc|cc|cc||cc|cc|cc@{}}
\toprule
 & \multicolumn{6}{c||}{\textbf{profession}} & \multicolumn{6}{c}{\textbf{gender}} \\
\cmidrule(lr){2-7} \cmidrule(lr){8-13}
\multirow{3}{*}{\textbf{Models}} & \multicolumn{2}{c|}{\textbf{MovieChAtt}} & \multicolumn{2}{c|}{\textbf{PersonaChat}} & \multicolumn{2}{c||}{\textbf{Reddit}} & \multicolumn{2}{c|}{\textbf{MovieChAtt}} & \multicolumn{2}{c|}{\textbf{PersonaChat}} & \multicolumn{2}{c}{\textbf{Reddit}} \\ 
\cmidrule(lr){2-3} \cmidrule(lr){4-5} \cmidrule(lr){6-7} \cmidrule(lr){8-9} \cmidrule(lr){10-11} \cmidrule(lr){12-13}
 & \multicolumn{1}{c}{MRR} & \multicolumn{1}{c|}{AU-} & \multicolumn{1}{c}{MRR} & \multicolumn{1}{c|}{AU-} & \multicolumn{1}{c}{MRR} & \multicolumn{1}{c||}{AU-} & \multirow{2}{*}{Acc} & \multicolumn{1}{c|}{AU-} & \multirow{2}{*}{Acc} & \multicolumn{1}{c|}{AU-} & \multirow{2}{*}{Acc} & \multicolumn{1}{c}{AU-} \\
 & micro / macro & \multicolumn{1}{c|}{ROC} & micro / macro & \multicolumn{1}{c|}{ROC} & micro / macro & \multicolumn{1}{c||}{ROC} &  & \multicolumn{1}{c|}{ROC} &  & \multicolumn{1}{c|}{ROC} &  & \multicolumn{1}{c}{ROC} \\
\midrule
Embedding sim.    & \sig{0.22} / \sig{0.14} & \sig{0.60} & \sig{0.30} / \sig{0.25} & \sig{0.63} & \sig{0.15} / \sig{0.13} & \sig{0.59} & \sig{0.52} & \sig{0.54} & \nsig{0.49} & \nsig{0.50} & \sig{0.61} & \sig{0.60} \\
Logistic reg.     & \sig{0.46} / \sig{0.20} & \sig{0.76} & \sig{0.81} / \sig{0.77} & \sig{0.58} & \sig{0.13} / \sig{0.19} & \nsig{0.57} & \nsig{0.59} & \nsig{0.62} & \nsig{0.86} & \nsig{0.93} & \sig{0.69} & \sig{0.75} \\ 
MLP               & \bnsig{0.47} / \nsig{0.20} & \nsig{0.75} & \sig{0.86} / \sig{0.77} & \nsig{0.97} & \nsig{0.46} / \nsig{0.23} & \nsig{0.78} & \sig{0.57} & \sig{0.60} & \nsig{0.80} & \nsig{0.87} & \nsig{0.71} & \nsig{0.77} \\
\midrule
N-GrAM \cite{basile:2017} & \sig{0.21} / \sig{0.16} & \sig{0.62} & \sig{0.83} / \nsig{0.83} & \nsig{0.88} & \nsig{0.17} / \nsig{0.26} & \sig{0.64} & \nsig{0.57} & \nsig{0.58} & \nsig{0.86} & \nsig{0.87} & \sig{0.66} & \sig{0.71} \\
W2V-C \cite{pietro:ACL15} & \sig{0.25} / \sig{0.13} & \sig{0.74} & \nsig{0.59} / \nsig{0.46} & \nsig{0.89} & \sig{0.27} / \sig{0.17} & \sig{0.74} & \nsig{0.62} & \nsig{0.66} & \sig{0.73} & \sig{0.80} & \sig{0.64} & \sig{0.73} \\
CNN \cite{bayot:MOD17} & \sig{0.19} / \sig{0.20} & \sig{0.66} & \sig{0.77} / \sig{0.77} & \sig{0.81} & \sig{0.26} / \sig{0.24} & \sig{0.76} & \nsig{0.60} & \nsig{0.60} & \sig{0.72} & \sig{0.73} & \sig{0.61} & \sig{0.61} \\
\midrule
\method{avg}      & \sig{0.39} / \sig{0.37} & \sig{0.81} & \sig{0.86} / \sig{0.91} & \sig{0.98} & \sig{0.34} / \sig{0.22} & \sig{0.82} & \nsig{0.72} & \nsig{0.82} & \nsig{0.79} & \nsig{0.87} & \bnsig{0.86} & \nsig{0.92} \\
\method{CNN}      & \nsig{0.42} / \nsig{0.37} & \nsig{0.83} & \bnsig{0.96} / \bnsig{0.94} & \bnsig{0.99} & \sig{0.36} / \sig{0.37} & \sig{0.86} & \nsig{0.75} & \bnsig{0.85} & \nsig{0.95} & \bnsig{0.99} & \bnsig{0.86} & \sig{0.93} \\ 
\hdashline
\method{CNN-attn} & \nsig{0.43} / \bnsig{0.50} & \bnsig{0.85} & \nsig{0.90} / \nsig{0.93} & \bnsig{0.99} & \bnsig{0.51} / \nsig{0.40} & \bnsig{0.9} & \bnsig{0.77} & \nsig{0.84} & \bnsig{0.96}  & \nsig{0.97}  & \nsig{0.85} & \bnsig{0.94} \\
\method{2attn}    & \nsig{0.39} / \nsig{0.34} & \nsig{0.84} & \nsig{0.94} / \nsig{0.93} & \bnsig{0.99} & \nsig{0.43} / \bnsig{0.42} & \nsig{0.89} & \nsig{0.69} & \nsig{0.77} & \nsig{0.94} & \nsig{0.98} & \nsig{0.80} & \nsig{0.91} \\
\bottomrule
\end{tabular}
\caption{Comparison of 
models on all datasets for \emph{profession} and \emph{gender} attributes. 
Results marked with * significantly differ from
the best method (in bold face)
with $p < 0.05$ as measured by a paired t-test (MRR) or McNemar's test (Acc and AUROC).}
\label{tab:model-comparison-profession-gender}
\end{table*}

\section{Results and Discussion}
\label{results}

\subsection{Main Findings}
\label{findings}
\www{
In Table \ref{tab:model-comparison-profession-gender}, \ref{tab:model-comparison-age} and \ref{tab:model-comparison-family} we report results for the HAMs and the baselines on all datasets (MovieChAtt, PersonaChat and Reddit) for all considered attributes (\textit{profession}, \textit{gender}, \textit{age} and \textit{family status}).
We do not report results for the \textit{pattern oracle} baseline in the tables as we evaluated the baseline solely on the MovieChAtt dataset, because the oracle essentially replicates the way we labeled persona descriptions and posts in the PersonaChat and Reddit datasets, respectively.
The pattern oracle baseline yields 0.21/0.20 micro/macro MRR for \textit{profession}, 0.67 accuracy for \textit{gender}, and 0.41/0.40 micro/macro MRR for \textit{age}. HAMs significantly outperform this baseline, indicating that identifying explicit mentions of attribute values is insufficient in our dialogue setting.

HAMs outperform the distantly supervised models (i.e., \textit{embedding similarity}, \textit{logistic regression} and \textit{Multilayer Perceptron (MLP)}) in the vast majority of cases.
MLP and logistic regression perform best in several occasions for \textit{profession} and \textit{age} attributes when micro MRR is considered. However, their macro MRR scores fall behind HAMs, showing that HAMs are better at dealing with 
multi-valued attributes having skewed distribution.
The low performance of these distantly supervised methods may be related to their strong assumption that every sequence of four utterances contains information about the attribute being predicted.
}

\www{
Comparing with baselines from prior work, HAMs significantly outperform N-GrAM \cite{basile:2017} in many cases,
suggesting that representing utterances using word embeddings, instead of merely character and word n-grams, is important for this task. 
Using neural clusters (W2V-C) as features for the classification task \cite{pietro:ACL15} works quite well for the \textit{age} attribute, where different `topics' may correlate with different age categories (e.g. `video game' for \textit{teenager} and `office' for \textit{adult}).
However, W2V-C is often significantly worse for the \textit{profession}, \textit{gender}, and \textit{family status} attributes, which may be caused by similar discriminative words (e.g., `husband'/`wife' for \textit{gender}) being clustered together in the same topic.
The CNN baseline \cite{bayot:MOD17} is significantly worse than the best method in the majority of cases. Furthermore, it generally performs substantially worse than \method{CNN}, further illustrating the advantage of aggregating utterances within the model.
}

\www{
In general, \method{avg} performs worse than the other HAMs\footnote{
For the sake of brevity we neither instantiate nor report results for LSTM-based HAMs, such as $f_{utter}=\text{\textit{LSTM}}$ and $f_{subj}=attn{\text -}avg$ or $f_{subj}=\text{\textit{LSTM}}$. These models were unable to outperform \method{avg}, with the best variant obtaining a micro MRR of only 0.31 after grid search (profession predicate on MovieChAtt; Table \ref{tab:model-comparison-profession-gender}). This is in line with recent results suggesting that RNNs are not ideal for identifying semantic features \cite{tang2018self}. 
}, demonstrating that simple averaging is insufficient
for representing utterances and subjects.
In most cases \method{CNN} performs slightly worse than \method{CNN-attn}, demonstrating
the value of exploiting an attention mechanism to combine subject's utterances.
\method{CNN-attn} and \method{2attn}  
achieve the strongest performance across predicates, with \method{CNN-attn} generally performing better.
\method{CNN-attn} performs particularly well on the \textit{gender} and \textit{family status} attributes, where detecting bigrams may yield an advantage. For example, \method{2attn} places high attention weights on terms like `family' and `girlfriend' where the previous term may be a useful signal (e.g., `my family' vs. `that family').

The gap between the baselines and HAMs is often smaller on PersonaChat compared with on the other two datasets, illustrating the simplicity of crowdsourced-dialogue as compared to movie scripts or Reddit discussions. This is also supported by the fact that the maximum metrics on PersonaChat are much higher. There are several factors that may be responsible for this: \textit{(1)} the dialogue in PersonaChat was created by crowdworkers with the goal of stating facts that were given to them, which often leads to the facts being stated in a straightforward manner (e.g., ``I am an author''); \textit{(2)} PersonaChat utterances are much shorter 
and there are far fewer utterances per character (i.e., a maximum of 4 in PersonaChat vs. a minimum of 20 in MovieChAtt), leading to a higher density of information related to attributes; and \textit{(3)} the persona descriptions in PersonaChat are used for many dialogues, giving models an opportunity to learn specific personas.
}

\begin{table}[t]
\centering
\small
\begin{tabular}{@{}lcc|cc@{}}
\toprule
& \multicolumn{4}{c}{\textbf{age}} \\
\cmidrule(lr){2-5}
\multirow{3}{*}{\textbf{Models}} & \multicolumn{2}{c|}{\textbf{MovieChAtt}} & \multicolumn{2}{c}{\textbf{Reddit}} \\ 
\cmidrule(lr){2-3} \cmidrule(lr){4-5} 
 & \multicolumn{1}{c}{MRR} & \multicolumn{1}{c|}{AU-} & \multicolumn{1}{c}{MRR} & \multicolumn{1}{c}{AU-} \\
 & micro / macro & \multicolumn{1}{c|}{ROC} & micro / macro & \multicolumn{1}{c}{ROC} \\
\midrule
Embedding sim.    & \sig{0.45} / \sig{0.45} & \sig{0.61} & \sig{0.55} / \sig{0.44} & \sig{0.56} \\
Logistic reg.     & \sig{0.65} / \sig{0.49} & \nsig{0.76} & \bnsig{0.80} / \nsig{0.61} & \nsig{0.87} \\ 
MLP               & \sig{0.64} / \sig{0.48} & \nsig{0.83} & \nsig{0.78} / \nsig{0.48} & \nsig{0.88} \\
\midrule
N-GrAM \cite{basile:2017} & \nsig{0.69} / \nsig{0.47} & \nsig{0.85} & \sig{0.48} / \sig{0.53} & \sig{0.55} \\
W2V-C \cite{pietro:ACL15} & \nsig{0.67} / \nsig{0.45} & \nsig{0.86} & \nsig{0.75} / \nsig{0.51} & \nsig{0.88} \\
CNN \cite{bayot:MOD17} & \sig{0.66} / \sig{0.62} & \nsig{0.83} & \sig{0.68} / \sig{0.65} & \sig{0.79} \\
\midrule
\method{avg}      & \sig{0.62} / \nsig{0.59} & \sig{0.76} & \nsig{0.67} / \nsig{0.67} & \sig{0.77} \\
\method{CNN}      & \sig{0.73} / \bnsig{0.63} & \nsig{0.84} & \sig{0.73} / \sig{0.61} & \sig{0.89} \\ 
\hdashline
\method{CNN-attn} & \nsig{0.73} / \nsig{0.60} & \bnsig{0.86} & \nsig{0.79} / \bnsig{0.68} & \bnsig{0.90} \\
\method{2attn}    & \bnsig{0.74} / \nsig{0.6} & \nsig{0.85} & \nsig{0.72} / \nsig{0.6} & \nsig{0.82} \\
\bottomrule
\end{tabular}
\caption{Comparison of 
models on all datasets for \emph{age} attribute. 
Results marked with * significantly differ from 
the best method (in bold face)
with $p < 0.05$ as measured by a paired t-test (MRR) or McNemar's test (Acc and AUROC).}
\label{tab:model-comparison-age}
\end{table}

\begin{table}[t]
\centering
\small
\begin{tabular}{@{}lcc|cc@{}}
\toprule
& \multicolumn{4}{c}{\textbf{family status}} \\
\cmidrule(lr){2-5}
\multirow{2}{*}{\textbf{Models}} & \multicolumn{2}{c|}{\textbf{PersonaChat}} & \multicolumn{2}{c}{\textbf{Reddit}} \\ 
\cmidrule(lr){2-3} \cmidrule(lr){4-5} 
 & Acc & AUROC & Acc & AUROC \\
\midrule
Embedding sim.    & \sig{0.41} & \sig{0.49} & \sig{0.42} & \sig{0.47} \\
Logistic reg.     & \sig{0.75} & \sig{0.84} & \nsig{0.71} & \nsig{0.74} \\ 
MLP               & \nsig{0.70} & \nsig{0.80} & \sig{0.62} & \sig{0.60} \\
\midrule
N-GrAM \cite{basile:2017} & \nsig{0.85} & \nsig{0.86} & \sig{0.45} & \sig{0.47} \\
W2V-C \cite{pietro:ACL15} & \sig{0.74} & \sig{0.82} & \nsig{0.70} & \nsig{0.78} \\
CNN \cite{bayot:MOD17} & \nsig{0.74} & \nsig{0.74} & \nsig{0.69} & \nsig{0.69} \\
\midrule
\method{avg}      & \nsig{0.80} & \nsig{0.91} & \nsig{0.67} & \nsig{0.72} \\
\method{CNN}      & \bnsig{0.93} & \bnsig{0.99} & \sig{0.52} & \sig{0.62} \\ 
\hdashline
\method{CNN-attn} & \nsig{0.92} & \nsig{0.98} & \bnsig{0.70} & \bnsig{0.78} \\
\method{2attn}    & \nsig{0.88} & \nsig{0.94} & \nsig{0.64} & \nsig{0.67} \\
\bottomrule
\end{tabular}
\caption{Comparison of 
models on all datasets for \emph{family status} attribute. 
Results marked with * significantly differ from 
the best method (in bold face)
with $p < 0.05$ as measured by a paired t-test (MRR) or McNemar's test (Acc and AUROC).}
\label{tab:model-comparison-family}
\end{table}

\www{
\subsection{Study on word embeddings}
In the previous experiments, we represented terms using embeddings from a word2vec skip-gram model trained on Google News. \cite{embed1}
In this study we compare the Google News embeddings with word2vec embeddings trained on Reddit posts, GloVe \cite{pennington2014glove} embeddings trained on Common Crawl, and GloVe embeddings trained on Twitter. 
We also consider $ELMo$ \cite{Peters:2018}, a contextualized embedding model. To capture semantic variations, this model creates a contextualized character-based representation of words using a bidirectional language model. We use AllenNLP's small ELMo model\footnote{https://allennlp.org/elmo} trained on the 1 Billion Word Benchmark of news crawl data from WMT 2011 \cite{41880}. 

Given space limitations and the higher model variance on the \textit{profession} attribute on MovieChAtt, we restrict the study to this predicate and corpus. We evaluated the two best performing HAMs, i.e., \method{CNN-attn} and \method{2attn}. Table \ref{emb_tab} shows the results obtained with the various embedding methods trained on different datasets.
The difference in performance does not greatly vary across embedding models and datasets, with Google News embeddings performing best in terms of macro MRR and Reddit embeddings performing best in terms of micro MRR.
Despite their strong performance on some NLP tasks, the ELMo contextualized embeddings do not yield a performance boost for any method or metric.
We view this observation as an indicator that the choice of term embedding method is not very significant for this task compared to the method used to combine terms into an utterance representation.
}

\begin{table}[]
\centering
\small
\begin{tabular}{@{}l@{\hskip 1\tabcolsep}l@{}c@{\hskip 1\tabcolsep}cc@{\hskip 1\tabcolsep}c@{}}
\toprule
\multirow{3}{*}{\textbf{Model}} & \multirow{3}{*}{\textbf{Corpus}} & \multicolumn{2}{c}{\textbf{\method{CNN-attn}}} & \multicolumn{2}{c}{\textbf{\method{2attn}}} \\
 \cmidrule(lr){3-4} \cmidrule(lr){5-6}
 &  & MRR & AU- & MRR & AU- \\
 &  & micro / macro & ROC & micro / macro & ROC\\
 \midrule
word2vec & Google News & 0.42 / \textbf{0.44} & 0.77 & 0.39 / 0.37 & \textbf{0.83} \\
(skip-gram) & Reddit & \textbf{0.43} / 0.37 & \textbf{0.82} & \textbf{0.50} / 0.37 & \textbf{0.83} \\
 \midrule
 \multirow{2}{*}{GloVe} & Common Crawl & 0.40 / 0.37 & 0.76 & 0.40 / \textbf{0.39} & 0.82 \\
 & Twitter & 0.39 / 0.35 & 0.67 & 0.36 / 0.34 & 0.81 \\
 \midrule
 ELMo & WMT News & 0.38 / 0.32 & 0.76  & 0.37 / 0.37 & 0.83 \\
\bottomrule
\end{tabular}
\caption{Comparison of embedding models trained on different datasets, for identifying \textit{profession} attribute.
}
\label{emb_tab}
\end{table}

\subsection{Ablation study}

We performed an ablation study in order to determine the performance impact of the HAMs' components.
Given space limitations and the higher model variance on the \textit{profession} attribute on MovieChAtt, we restrict the study to this predicate and dataset. 
Ablation results for \method{2attn} using cross validation on the training set are shown in Table \ref{tab4}.
Replacing either representation function (i.e., $f_{utter}$ or $f_{subj}$) with an averaging operation reduces performance, as shown in the last two lines. Attention on utterance representations ($R^{utter}$) is slightly more important in terms of MRR, but both types of attention contribute to \method{2attn}'s performance.
Similarly, removing both types of attention corresponds to \method{avg}, which consistently underperforms \method{2attn} in Table \ref{tab:model-comparison-profession-gender}, \ref{tab:model-comparison-age} and \ref{tab:model-comparison-family}.

Removing attention from \method{CNN-attn} yields \method{CNN}, which consistently performs worse than \method{CNN-attn} in Table \ref{tab:model-comparison-profession-gender}, \ref{tab:model-comparison-age} and \ref{tab:model-comparison-family},
supporting the observation that attention is important for performance on our task.
Intuitively, attention provides a useful signal because it allows the model to focus on only those terms that provide information about an attribute.

\begin{figure*}[th!]
 \centering
 \begin{subfigure}{.9\textwidth}
   \centering
   \includegraphics[scale=0.43]{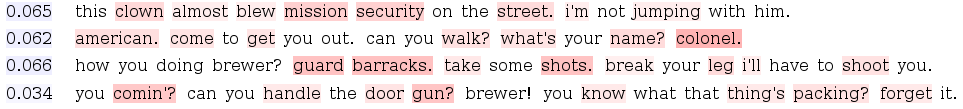}
   \caption{\textit{profession}: military personnel}
   \label{fig:att-profession}
 \end{subfigure}
 \\[8pt]
 \begin{subfigure}{.9\textwidth}
   \centering
   \includegraphics[scale=0.43]{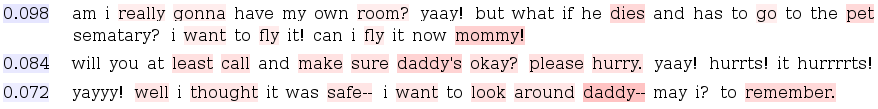}
   \caption{\textit{age} (category): child}
   \label{fig:att-age}
 \end{subfigure}
\vspace*{-0.3cm}
\caption{Attention visualization for \textit{profession} and \textit{age} predicates on MovieChAtt.}
\label{fig:att-profession-age}
\end{figure*}

\begin{figure*}[th!]
 \centering
 \begin{subfigure}{.9\textwidth}
   \centering
   \includegraphics[scale=0.17]{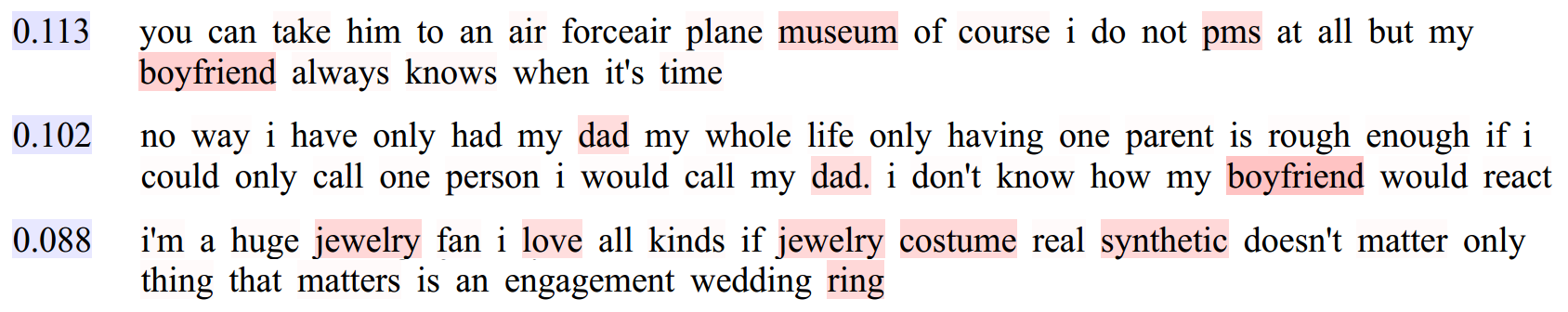}
   \caption{\textit{gender}: female}
   \label{fig:att-gender}
 \end{subfigure}
 \\[8pt]
 \begin{subfigure}{.9\textwidth}
   \centering
   \includegraphics[scale=0.17]{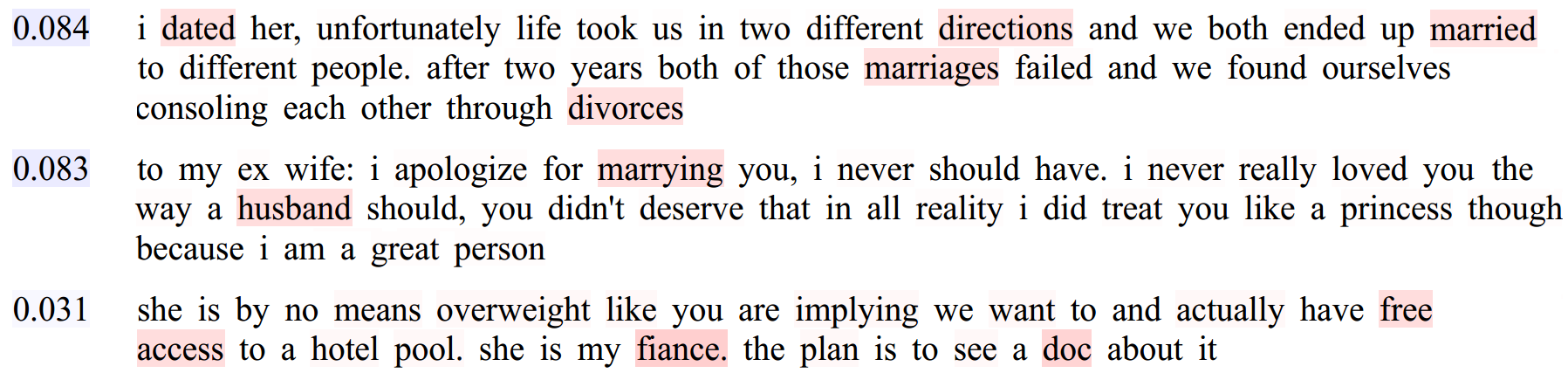}
   \caption{\textit{family status}: married}
   \label{fig:att-family}
 \end{subfigure}
\vspace*{-0.3cm}
\caption{Attention visualization for \textit{gender} and \textit{family status} predicates on Reddit.}
\label{fig:att-gender-family}
\end{figure*}

\commented{
\begin{figure*}[th!]
\centering
  \centering
  \includegraphics[scale=0.6]{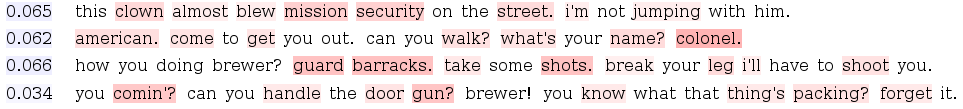}
  \caption{Attention visualization for profession `military personnel'}
  \label{fig:att-a}
\end{figure*}%

\begin{figure*}[th!]
  \centering
  \includegraphics[scale=0.6]{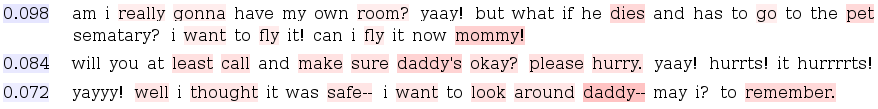}
  \caption{Attention visualization for age category `child'}
  \label{fig:att-b}
\end{figure*}

\begin{figure*}[th!]
  \centering
  \includegraphics[scale=0.8]{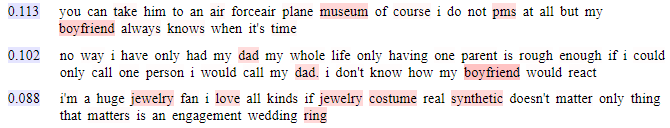}
  \caption{Attention visualization for gender `female'}
  \label{fig:att-b}
\end{figure*}

\begin{figure*}[th!]
  \centering
  \includegraphics[scale=0.8]{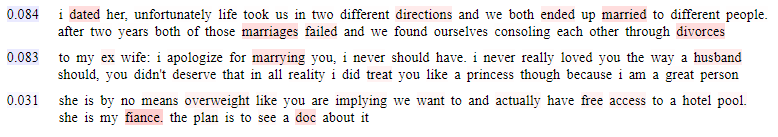}
  \caption{Attention visualization for family status `married'}
  \label{fig:att-b}
\end{figure*}
}

\begin{table}[t!]
\centering
\small
\begin{tabular}{@{}lcc@{}}
\toprule
                         & \multicolumn{1}{c}{MRR} & \multirow{2}{*}{AUROC} \\
                         & micro / macro &  \\
                         \midrule
\method{2attn}           & \textbf{0.57} / \textbf{0.42} & \textbf{0.84} \\
$-$ attention on terms       & 0.49 / 0.40 & 0.81 \\
$-$ attention on $R^{utter}$ & 0.48 / 0.34 & 0.82 \\
\bottomrule
\end{tabular}
\caption{Ablation study for the \textit{profession} attribute.}
\label{tab4}
\end{table}

\subsection{Case study on attention weights}
\www{
In order to illustrate the types of terms the model is looking for, we display \method{2attn}'s term and utterance weights for 
\textit{profession} and \textit{age} attributes (on MovieChAtt) in Figure \ref{fig:att-profession-age}, as well as \textit{gender} and \textit{family status} attributes (on Reddit) in Figure \ref{fig:att-gender-family}.
While \method{2attn} is sometimes outperformed by \method{CNN-attn}, this model is more interpretable because individual terms are considered in isolation.
Note that all these dialogue snippets come from the specific
datasets in our experiments, partly from fictional
conversations.
Some are skewed and biased, and not representative
for the envisioned downstream applications.

When predicting \textit{military personnel} as the \textit{profession} (Figure \ref{fig:att-profession}), the model focuses on military terms such as \textit{mission}, \textit{guard}, \textit{barracks}, and \textit{colonel}.
When predicting \textit{child} as the \textit{age category} (Figure \ref{fig:att-age}), on the other hand, the model focuses on terms a child is likely to use, such as \textit{pet}, \textit{mommy}, and \textit{daddy}.
According to Reddit posts, \textit{female} gender is suggested by terms such as \textit{boyfriend}, \textit{pms} and \textit{jewelry} (Figure \ref{fig:att-gender}). Meanwhile, \textit{married} users were identified through terms such as \textit{dated}, \textit{fiance} and \textit{divorces}, along with obvious terms like \textit{marrying} and \textit{marriages} (Figure \ref{fig:att-family}). 
These examples illustrate how the model is able to infer a subject's attribute by aggregating signals across utterances.
}

In addition to looking at specific utterances, we investigated which terms the model is strongly associating with a specific attribute. To do so, we computed attribute value probabilities for each term in the corpus, and kept the top terms for each attribute value. The results using \method{2attn} are shown in Table \ref{tab6}, which is divided into words that appear informative (top section) and words that do not (bottom section). In the case of informative words, there is a clear relationship between the words and the corresponding profession. 
Many of the uninformative words appear to be movie-specific, such as names (e.g., xavier, leonard) and terms related to a \textit{waiter's} role in a specific movie (e.g., rape, stalkers). Reducing the impact of setting-specific signals like this is one direction for future work.

\begin{table}[t]
\centering
\small
\begin{tabular}{@{}l@{\hskip 0.5\tabcolsep}l@{}}
\toprule
\textbf{profession} & \textbf{significant words} \\
\midrule
\textit{scientist} & characteristics, theory, mathematical, species, changes \\
\textit{politician} & governors, senate, secretary, reporters, president \\
\textit{detective} & motel, spotted, van, suitcase, parked \\
\textit{military personnel} & captured, firepower, guard, soldiers, attack \\
\midrule
\textit{student} & playing, really, emotional, definitely, unbelievable \\
\textit{photographer} & xavier, leonard, collins, cockatoo, burke \\
\textit{waiter} & rape, stalkers, murdered, overheard, bothering \\
\bottomrule
\end{tabular}
\caption{Top-5 words from \method{2attn} characterizing each profession.
}
\label{tab6}
\end{table}

\begin{table}[t]
\centering
\small
\begin{tabular}{@{}lc@{}cc@{\hskip 1\tabcolsep}cc@{}c@{}}
\toprule
\multirow{3}{*}{\textbf{Models}} & \multicolumn{2}{c}{\textbf{profession}\vspace{3pt}} & \multicolumn{2}{c}{\textbf{gender}} & \multicolumn{2}{c}{\textbf{age}} \\
\cmidrule(lr){2-3} \cmidrule(lr){4-5} \cmidrule(lr){6-7}
 & \multicolumn{1}{c}{MRR} & AU- & \multicolumn{1}{c}{\multirow{2}{*}{Acc}} & AU-  & \multicolumn{1}{c}{MRR} & AU-  \\
 & \multicolumn{1}{c}{micro / macro} & ROC &  & ROC & \multicolumn{1}{c}{micro / macro} & ROC \\
\midrule
\Tstrut \method{CNN-attn}     & 0.19 / 0.18 & 0.58 & 0.56 & 0.58 & \textbf{0.57} / \textbf{0.54} & \textbf{0.69} \\ 
\method{2attn}                & \textbf{0.21} / \textbf{0.21} & \textbf{0.67} & \textbf{0.61} & \textbf{0.64} & 0.45 / 0.41 & 0.45 \\
\bottomrule
\end{tabular}
\caption{Transfer learning performance of pre-trained Reddit models on MovieChAtt.}
\label{tab:transfer-learning-reddit-moviechatt}
\end{table}

\vspace{-5pt}
\subsection{Insights on transfer learning}

\www{
To 
investigate the robustness of our trained HAMs,
we tested 
the best performing models (i.e., \method{2attn} and \method{CNN-attn})
on a transfer learning task between our datasets. 
Specifically, we leveraged user-generated social media text (Reddit) available in abundance for inferring personal attributes of subjects in speech-based dialogues (MovieChAtt and PersonaChat). 
We report the results in Table \ref{tab:transfer-learning-reddit-moviechatt} and \ref{tab:transfer-learning-reddit-personachat} respectively.

While the scores on PersonaChat are low compared to those in Table \ref{tab:model-comparison-profession-gender} and \ref{tab:model-comparison-family}, the HAMs' performance on MovieChAtt is often comparable with the baselines' performance in Table \ref{tab:model-comparison-profession-gender}, \ref{tab:model-comparison-age} and \ref{tab:model-comparison-family}.
This difference may be caused by the fact that PersonaChat is a smaller, more synthetic dataset as discussed in Section \ref{findings}.
On MovieChAtt with the \textit{profession} predicate, both HAMs match the performance of all six baselines in terms of macro MRR. Similarly, \method{2attn} matches the performance of five of the six baselines on the \textit{gender} predicate (accuracy), and \method{CNN-attn} matches the performance of four of the six baselines on the \textit{age} predicate (macro MRR).
Particularly for the \textit{profession} attribute, missing training subjects in the Reddit dataset for certain professions such as \textit{astronaut} or \textit{monarch} contribute to the decreasing performance, although the models still make a reasonable prediction of \textit{scientist} for \textit{astronaut}. 
The methods do not perform as well in terms of micro MRR, which may be due to the substantially different attribute value distributions between datasets (i.e., the professions common in movies are uncommon in Reddit).
Improving the HAMs' transfer learning performance is a direction for future work.
}

\begin{table}[t]
\centering
\small
\begin{tabular}{@{}lcccccc@{}}
\toprule
\multirow{3}{*}{\textbf{Models}} & \multicolumn{2}{c}{\textbf{profession}\vspace{3pt}} & \multicolumn{2}{c}{\textbf{gender}} & \multicolumn{2}{c}{\textbf{family status}} \\
\cmidrule(lr){2-3} \cmidrule(lr){4-5} \cmidrule(lr){6-7} 
 & \multicolumn{1}{c}{MRR} & AU- & \multicolumn{1}{c}{\multirow{2}{*}{Acc}} & AU-  & \multicolumn{1}{c}{\multirow{2}{*}{Acc}} & AU- \\
 & \multicolumn{1}{c}{micro / macro} & ROC &  & ROC &  & ROC \\
\midrule
\Tstrut \method{CNN-attn}     & 0.20 / 0.16 & 0.58 & \textbf{0.52} & 0.50 & \textbf{0.74} & \textbf{0.74} \\
\method{2attn}                & \textbf{0.21} / \textbf{0.18} & \textbf{0.71} & 0.51 & \textbf{0.54} & 0.62 & 0.64 \\
\bottomrule
\end{tabular}
\caption{Transfer learning performance of pre-trained Reddit models on PersonaChat.}
\label{tab:transfer-learning-reddit-personachat}
\end{table}

\subsection{Profession misclassification study}
\begin{figure}[t]
\centering
\includegraphics[width=0.45\textwidth]{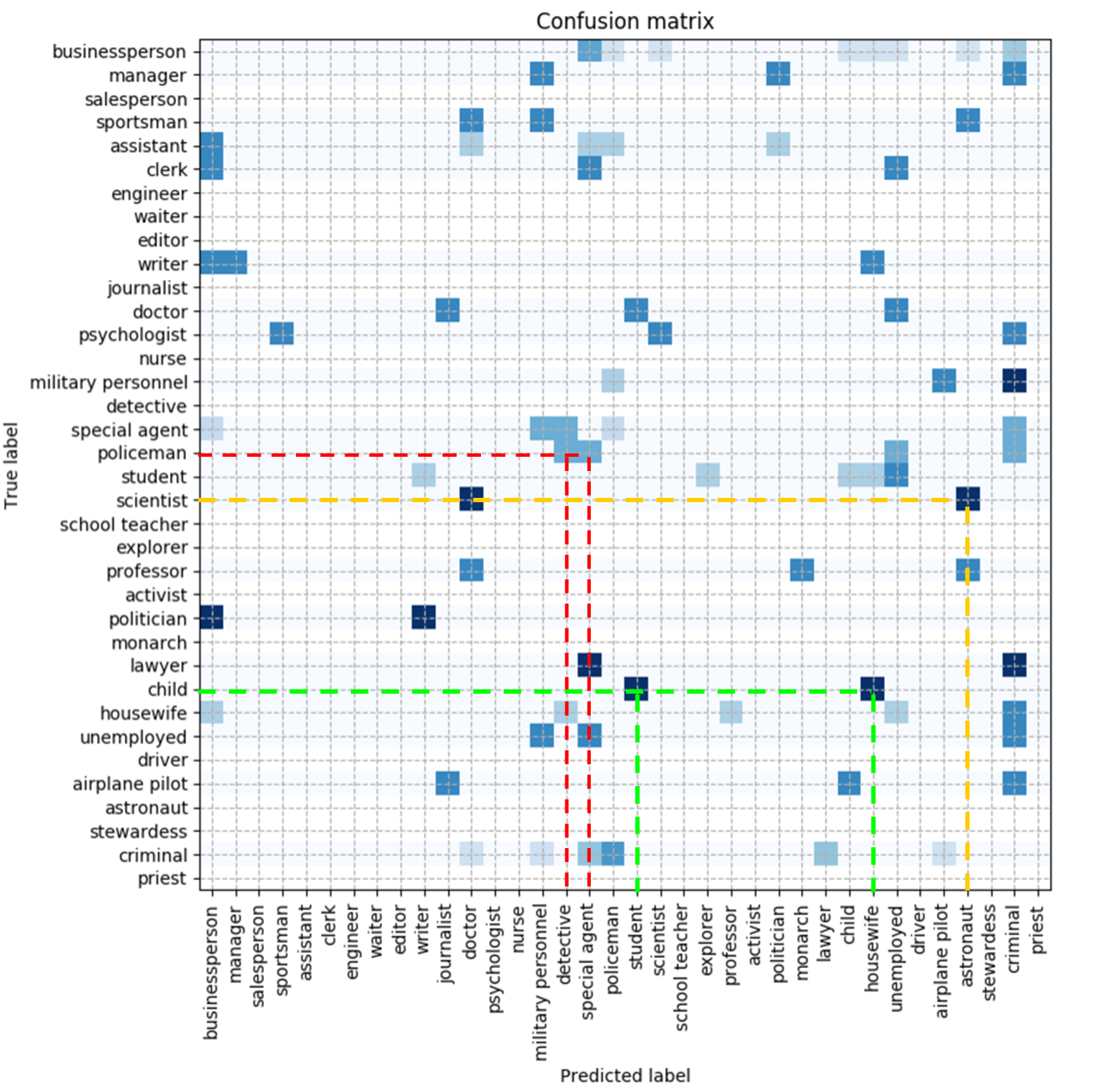}
\vspace*{-0.3cm}
\caption{Confusion matrix computed with \method{2attn}. True positives are not shown. Darker squares indicate more misclassifications.}
\label{matrix}
\end{figure}

In this section we investigate common misclassifications on the MovieChAtt dataset for the \textit{profession} predicate, which is both the most challenging predicate and the predicate with the most possible object values (i.e., 43 on MovieChAtt).
A confusion matrix for \method{2attn} is shown in Figure \ref{matrix}.
Dotted lines indicate several
interesting misclassifications: \textit{policemen} are often confused with \textit{detectives} and \textit{special agents} (red line); \textit{scientists} are confused with \textit{astronauts} (yellow line), because sci-fi films often feature characters who arguably serve in both roles; and a \textit{child} is often labeled as a \textit{student} or a \textit{housewife} (green line) because they sometimes use similar terms (e.g., `school' is used by both children and students, and `mommy' is used by both children and housewives).
Finally, 
many occupations are confused with \textit{criminal}, which is the most common profession in MovieChAtt.

\www{
To further compare the performance of the model on direct and transfer learning tasks we computed the confusion matrix for \method{2attn} trained on the Reddit dataset and using MovieChAtt as the test corpus. Interesting misclassifications include the following: artistic professions (\textit{actor}, \textit{painter}, \textit{musician}, \textit{director}) are often mixed up (red lines); \textit{banker} is confused with \textit{manager} (green lines); \textit{policeman} and \textit{airplane pilot} are confused with \textit{military personnel} (purple lines); and \textit{stewardess} is often confused with \textit{nurse} as they both are related to caring and serving tasks (yellow line).
}

\begin{figure}[t]
\centering
\includegraphics[width=0.34\textwidth]{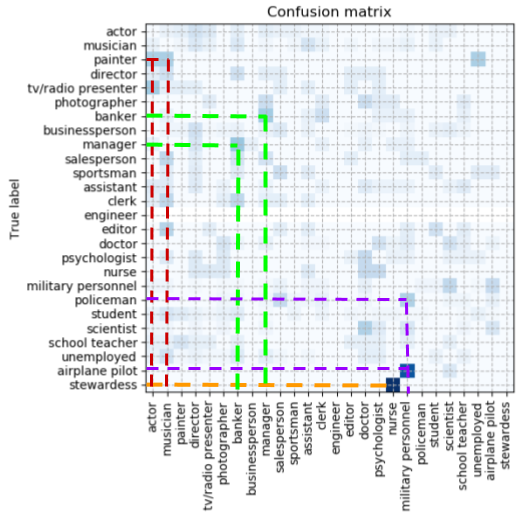}
\vspace*{-0.3cm}
\caption{Confusion matrix of Reddit$_{\text{train}}$ $\rightarrow$ MovieChAtt$_{\text{test}}$ computed with \method{2attn}. True positives are not shown. Darker squares indicate more misclassifications.}
\label{matrix-reddit}
\end{figure}

%% file: sections/conclusion.tex
\section{Conclusion}
We proposed Hidden Attribute Models (HAMs) for inferring personal attributes from conversations, such as a person's profession. 
We demonstrated the viability of our approach in extensive experiments considering several attributes on three datasets with diverse characteristics: Reddit discussions, movie script dialogues and crowdsourced conversations. 
We compared HAMs against a variety of state-of-the-art baselines, and achieved
substantial improvements over all of them, most notably by the \method{CNN-attn}
and \method{2attn} variants. 
We also demonstrated that the attention weights assigned by our methods provide
informative explanations of the computed output labels.

As a stress test for our methods, we investigated transfer learning by
training HAMs on one dataset and applying the learned models to other datasets.
Although we observed degradation in output quality, compared to
training on in-domain data, it is noteworthy that the transferred HAMs
matched the MovieChAtt performance of the baselines when trained on in-domain data.
We plan to further investigate this theme of transfer learning in our future work
on construction personal knowledge bases and leveraging them for personalized Web agents.